\documentclass[acmlarge]{acmart}

\AtBeginDocument{%
  }

\newtheorem{thm}{Theorem}
\newtheorem{cor}{Corollary}
\usepackage{amsmath}
\usepackage{amsmath}
\usepackage{hyperref}
\usepackage{tabularx}
\usepackage{color}
\usepackage{booktabs}
\usepackage{systeme}
\usepackage{mathtools}
\usepackage{amsfonts}
\usepackage{tikz}
\usepackage{pgfplots}
\usepackage{cellspace}
\usepackage{graphics}
\usepackage{subfig}
\usepackage{multirow}
\usepackage{algorithm}
\usepackage{diagbox}
\usepackage{enumitem}
\usepackage{cellspace}
\usepackage{color}
\usepackage{colortbl}
\usepackage{diagbox}
\usepackage{makecell}
\usepackage{multirow}
\usepackage{pifont}
\usepackage{tabu, booktabs}

\setcopyright{acmlicensed}
\copyrightyear{2024}
\acmYear{2024}

\begin{document}

\title{\Large Harnessing the Power of Noise: A Survey of Techniques and Applications}

\author{Reyhaneh Abdolazimi}
\email{rabdolaz@data.syr.edu}
\affiliation{%
  \institution{Data Lab, EECS Department, Syracuse University}
  \country{USA}
}

\author{Shengmin Jin}
\email{shengmij@amazon.com}
\affiliation{%
  \institution{Amazon}
  \country{USA}
  }

\author{Pramod K. Varshney}
\email{varshney@syr.edu}
\affiliation{%
  \institution{EECS Department, Syracuse University}
  \country{USA}
}

\author{Reza Zafarani}
\email{reza@data.syr.edu}
\affiliation{%
 \institution{Data Lab, EECS Department, Syracuse University}
 \country{USA}
 }

\renewcommand{\shortauthors}{Abdolazimi et al.}

\begin{abstract}
 Noise, traditionally considered a nuisance in computational systems, is reconsidered for its unexpected and counter-intuitive benefits across a wide spectrum of domains, including nonlinear information processing, signal processing, image processing, machine learning, network science, and natural language processing. Through a comprehensive review of both historical and contemporary research, this survey presents a dual perspective on noise, acknowledging its potential to both disrupt and enhance performance. Particularly, we highlight how noise-enhanced training strategies can lead to models that better generalize from noisy data, positioning noise not just as a challenge to overcome but as a strategic tool for improvement. This work calls for a shift in how we perceive noise, proposing that it can be a spark for innovation and advancement in the information era.
\end{abstract}



\maketitle

\section{Introduction}
In Computer science and across various engineering fields, noise is often considered a nuisance and annoyance. It distorts  details and makes data less accurate. In the past, the goal has often been to eliminate noise with the goal to make systems more reliable and accurate. But views on noise are changing. New findings suggest that noise can actually enhance and advance technologies in many areas, making us see it not just as a disruption but as a way to improve system performance.

Thus, once unwanted and hard to control, noise now appears to be a key player in improving the performance of complex information processing systems~\cite{chen2014noise}. This phenomena is often known as \textit{Stochastic Resonance}, which helps clear up signals, improve image quality, and strengthen models in machine learning~\cite{chen2014noise, osoba2013noise, audhkhasi2016noise}. This duality of noise --- both a problem and a benefit --- highlights the tricky role of noise while optimizing advanced neural networks and machine learning models.

Noise becomes even more notable when we consider real-world data, which is naturally noisy and often degrades the performance of trained neural network models in practical settings. Here lies an important contradiction: the same noisiness that causes system performance degradation also makes training datasets richer, by making networks more robust by helping them learn to apply their knowledge more broadly and effectively in real situations. This survey looks into the long history and exciting future of making use of noise in various domains. It discusses important and new studies that show how noise can be helpful in a variety of domain. By looking at how randomness can be a good thing—from improving signal detection systems and making learning algorithms stronger to protecting privacy and enhancing network science—we show how noise can have a positive impact and really change systems for the better.

The structure of this survey is carefully designed to guide researchers through the wide variety of methods and applications of noise enhancement. The first part, ``Fundamentals of Noise Benefits," begins with an exploration of the historical and foundational principles of noise enhancement that have shaped its use across various scientific fields. This section takes a closer look at noise in signal processing and image processing, which is segmented into areas such as signal detection—highlighting how noise can improve the ability to detect and interpret signals; parameter estimation—focusing on the impact of noise on the accuracy of estimated parameters; dithering—a technique vital for improving the quality of digital images by reducing the impact of quantization errors; and diverse applications of noise in enhancing image quality.

The second part of the survey, ``Noise in Machine Learning and Machine Learning Applications," explores how noise is used in unsupervised learning, discussing how noise can speed up the convergence of Markov chains and enhance the convergence in Expectation Maximization (EM) algorithms. This section also covers the positive role of noise in deep learning, detailing methods that utilize noise to boost training and generalization in neural networks. This includes noise injection strategies for inputs, outputs, hidden layers, network weights, and gradients. This part also examines how noise positively impacts machine learning applications, specifically in areas like graph machine learning, natural language processing, and recommendation systems. It discusses how noise can enhance graph connectivity, community detection, link prediction, graph augmentations, and strengthen language models to better manage real-world variations.

Several other applications of noise are explored in the third part. In this section, the discussion is broadened to include noise-enhanced quantum physics and neuroscience. This demonstrates that noise serves not only as a technical tool but also as a crucial component in advancing our understanding and capabilities in diverse fields such as quantum physics and biological systems. Additionally, the role of noise in enhancing data privacy and optimization is also discussed.

This survey challenges researchers to rethink the role of noise, not as something to simply remove, but as an essential tool in our scientific and technological toolbox. While there are a few surveys that focus on specific aspects of noise-enhanced systems, this survey is unique. It is the first to explore how noise impacts a broad range of fields—from signal processing and image processing to machine learning and network science—all together. This comprehensive approach shows the diverse applications of noise for system performance enhancement, encouraging us to exploit its potential to yield high quality real-world solutions.

\section{Fundamentals of Noise Benefits}
\noindent We begin with the history of noise enhancement and then provide an overview of engineered systems where noise helps in the enhancement of signal processing and image processing systems.

\subsection{Historical Foundations}

We first start with the history of this phenomenon, and in particular, \textit{Stochastic Resonance (SR)}. Roberto Benzi et al. first introduced the concept of SR in 1981 within the statistical physics community to explain the periodicity of Earth's ice ages~\cite{benzi1981mechanism}. As one of the initial applications of SR in signal processing, the authors of~\cite{gammaitoni1998stochastic} demonstrated how additive noise could improve the detectability of an input signal, as measured by the output signal-to-noise ratio (SNR). Over the years, SR has been applied to a variety of nonlinear systems where utilization of noise has enhanced the output signal quality, e.g.,~\cite{mcdonnell2009stochastic}. 

\begin{figure}[t]
	\centering
    \includegraphics[width=6 cm]{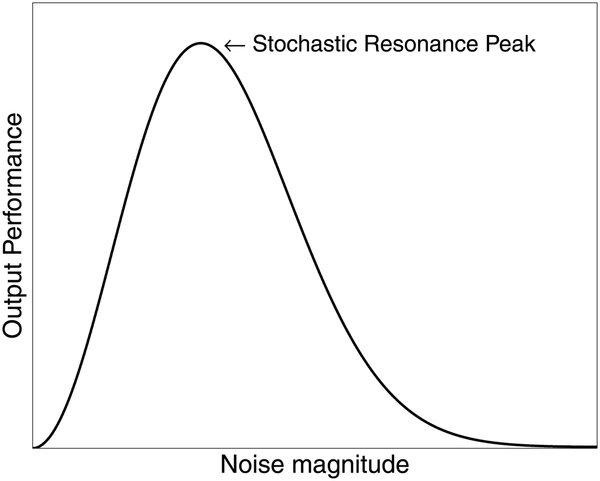}
    \caption{Typical curve of ``output performance" versus ``input noise magnitude" for systems capable of stochastic resonance~\cite{mcdonnell2009stochastic}. For small or large levels of noise, the performance improvement is limited, while performance reaches its maximum value when there is an appropriate amount of noise.}
    \label{fig:noiseperformance}
\end{figure}

Extensive reviews have focused on SR's mathematical formulations and applications in physical, electrical, and biological systems (see, for example, ~\cite{mcdonnell2009stochastic, Mcdonnellsuprathreshold,wiesenfeld1995stochastic, chen2014noise}). SR has been studied for various input signals, including periodic sine waves, periodic broadband signals, and aperiodic signals. Performance metrics used to evaluate the impact of SR include SNR~\cite{gammaitoni1998stochastic} that is suitable for periodic signals and broadband noise; Mutual Information (MI)~\cite{bulsara1996threshold} for random and aperiodic signals; and other metrics such as cross-correlation~\cite{collins1996aperiodic} and Fisher information~\cite{greenwood2000stochastic} for specific signal types. Research has shown that both low and excessive noise levels reduce system performance, whereas an optimal noise level significantly enhances it. Figure \ref{fig:noiseperformance} demonstrates how noise enhancement typically improves the performance of a nonlinear system. Performance, measured by metrics like the signal-to-noise ratio (SNR) or mutual information, is significantly lower at both small and large noise levels. However, an optimal, intermediate level of noise enhances performance substantially.

\subsection{Noise-Enhanced Signal Processing}

Numerous studies have explored the effect of noise enhancement in signal detection and estimation. 

\subsubsection{Signal Detection.} Utilization of noise plays a pivotal role in enhancing the detectability of signals in nonlinear detectors. For instance, noise can amplify a weak sinusoidal signal,  facilitating its detection~\cite{zozor2002use}. Similarly, usage of white Gaussian noise can significantly improve detector performance for identifying constant signals against a Gaussian mixture noise background under specific conditions~\cite{kay2000can}. Optimal detection performance is achieved when the noise parameters are finely tuned to match those of a suboptimal (locally optimal) detector~\cite{zozor2003stochastic}.

In a comprehensive series of studies, Chen et al. delve into enhancing detection performance via additive noise~\cite{chen2007theory, chen2008theory}. They introduce a mathematical framework that elucidates the stochastic resonance effect within binary hypothesis testing for both fixed and variable detectors~\cite{chen2007theory, chen2008theory}. We briefly summarize their findings. Consider observations $x \in X$ and binary hypotheses $H_0$ and $H_1$, with known probability density functions (pdfs) $p_0(x)$ and $p_1(x)$, and independent additive noise $\boldsymbol{n} \in N$ for enhancement with pdf $p_n(n) \in P_N$. This noise $\boldsymbol{n}$ can be added to input data $x$, yielding new data $y = x + \boldsymbol{n} \in Y$. The detector's performance can be evaluated using three metrics: (1) detection probability $P_d$, (2) false alarm rate $P_f$, and (3) error probability $P_e$. The strategies to enhance detection performance using noise for two detection frameworks are described as follows:

\begin{itemize}
    \item Under the \textit{Neyman–Pearson} criterion, maximization of the detection probability $P_d$ with the constraint $P_f \le \alpha$ involves the probabilistic addition of two discrete noise vectors: $$P_n^o = \lambda\delta(\boldsymbol{n}-\boldsymbol{n_1}) + (1-\lambda)\delta(\boldsymbol{n}-\boldsymbol{n_2}),$$ where $\delta$ represents the Dirac delta function, $\boldsymbol{n_1}$ and $\boldsymbol{n_2}$ are specific noise parameters, and $\lambda \in [0,1]$ is the randomization probability.
    
    \item Within the \textit{Bayesian} framework, minimization of error probability $P_e$ is achieved with a constant noise signal: $$P_{n,e}^o = \delta(\boldsymbol{n}-\boldsymbol{n_0}),$$ where $\boldsymbol{n_0}$ is a suitable noise parameter.
\end{itemize}

These principles apply to both fixed and variable detectors upon their concurrent adjustment with noise distribution selections~\cite{chen2008theory}. For more nuanced insights and theoretical justifications regarding enhancement in detectors via additive noise, readers are encouraged to consult~\cite{chen2007theory, chen2008theory, chen2014noise}.

Furthermore, sequential detection, where data collected sequentially is employed for detection, also benefits from addition of noise. This approach is shown to improve the detection performance of sequential detection systems by reducing the expected sample size needed for desired detection performance~\cite{chen2008improving}. The optimal noise for these circumstances can either be a constant signal or Gaussian noise with zero mean and variance $\sigma_n^2$. Additional examples and discussions on the constructive role of noise in signal detection can be found in~\cite{he2012effects, bayram2010noise, rousseau2005stochastic, rousseau2005constructive}

\subsubsection{Parameter Estimation.}
Noise injection into the input data can significantly enhance the performance of nonlinear estimators. Chen et al. explore a comprehensive framework for a noise-enhanced parameter estimator that incorporates both additive and nonadditive noise types~\cite{chen2008noise}. They analyze settings analogous to those in binary hypothesis detection, focusing on an input signal $x$, a parameter to be estimated $\theta$, and an estimator $\hat{\theta}=T(x)$ that derives an estimate of $\theta$ from $x$. With $y$ serving as the noisy input to the estimator, they model $n$ as an independent noise where $p(y \mid x,n) = \varsigma(x,n)$, and $\varsigma$ represents a predetermined stochastic transform function. Estimator performance is assessed using a risk function $r_i(\theta,\hat{\theta})$, with $i$ denoting specific performance metrics such as the Mean Square Error (MSE) or some statistics derived from the covariance matrix. They demonstrate that imposing constraints on the risk function, $r_i \le \alpha$, can be met by noise distributions of the form:
$$p_n(n) = \sum^{I}_{i=1} \lambda_i \delta(n-n_i),$$
where $\lambda_i \ge 0$, $\sum\lambda_i = 1$, and each $n_i$ corresponds to a specific constant vector (signal) relevant to the performance metrics.

Further results by Uhlich~\cite{uhlich2015bayes} demonstrate the impact of noise on reducing Bayes risk for estimators. By adding noise to estimators' observations and evaluating the expected output across a neighborhood of observations, Uhlich establishes conditions under which addition of noise can augment the estimator's efficacy. In a separate investigation into Bayes estimators, Duan et al. demonstrate how a linear mix of Bayes estimators, adjusted through variable weighting coefficients, benefits from noise to lower the Bayes risk for the MSE criterion, especially when the weighting coefficients across identical estimators align~\cite{duan2019noise}. While noise-enhanced signal processing harnesses stochastic resonance to improve the detectability of signals and estimation of parameters, a similar principle finds application in the realm of image processing. Here, noise is not just a signal inference booster but a quality enhancer, demonstrating the versatility of noise to enhance performance across different forms of data.

\subsection{Noise-Enhanced Image Processing}
Noise can help in improving various image processing techniques. We begin with a focus on \textit{dithering} and continue with various image processing applications that benefit from noise.
 
\subsubsection{Dithering.}
\begin{figure}[t]
	\centering
    \includegraphics[width=0.85\textwidth]{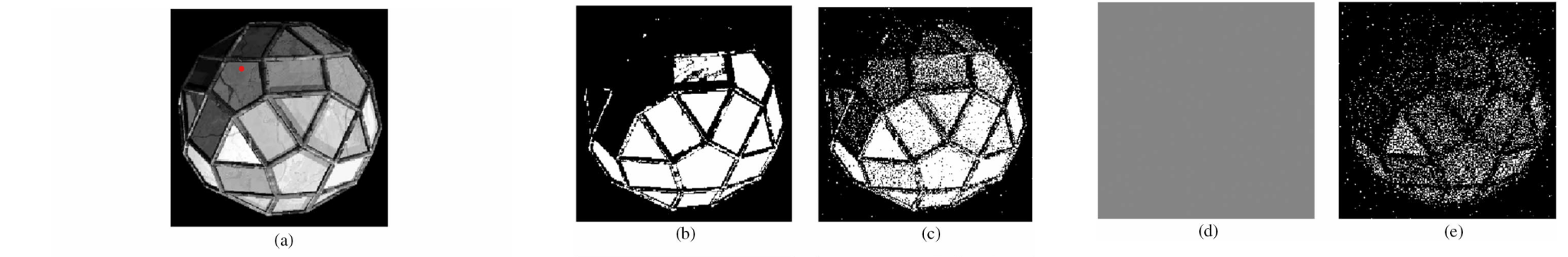}
    \caption{One-bit image quantization.~\cite{chen2014noise}.   (a) The original eight-bit image. (b) One-bit uniformly quantized image. (c) One-bit uniformly quantized
 image with additive noise. Noise helped to preserve details like contours and textures.  (d) One-bit subthreshold quantized image.
 (e) One-bit subthreshold quantized image with additive noise which helped to recover some visual information.}
    \label{fig:noiseimage}
\end{figure}

The concept of \textit{dithering} emerged in 1962 when Roberts demonstrated that adding random noise to images before quantization could minimize the difference between the input and output images, effectively randomizing quantization errors~\cite{roberts1962picture}. Originally applied to speech coding in the 1970s~\cite{jayant1972application}, dithering enhances static threshold nonlinear systems by introducing a random signal prior to digitization or quantization tasks~\cite{schuchman1964dither}. McDonnell et al.~\cite{Mcdonnellsuprathreshold} describe dithering as a technique that leverages stochastic resonance (SR), suggesting that the foundations of dithering and SR are not mutually exclusive. This led to investigations into how noise could improve human visual system performance, notably in signal detection and contrast sensitivity~\cite{sasaki2006effect}, and even in enhancing the recognition sensitivity of noisy images~\cite{piana2000role,simonotto1997visual}. An illustrative example of a basic image quantization task is shown in Figure \ref{fig:noiseimage}~\cite{chen2014noise}.  With a uniform quantizer, important details like contours and textures are lost, as seen in Figure \ref{fig:noiseimage}(b). However, when independent Gaussian noise is added before quantization, as in Figure \ref{fig:noiseimage}(c), these details are better preserved. This concept is also useful for detecting subthreshold signals: without noise, the image may disappear entirely (Figure \ref{fig:noiseimage}(d)), but adding noise can recover some visual information, as shown in Figure \ref{fig:noiseimage}(e). Additional details on dithering's impact on signal quality in quantization can be found in~\cite{720541, wannamaker2000stochastic,wannamaker2000theory}.\vspace{1mm}

\subsubsection{Noise Enhanced Image Processing Applications}
Enhancement via noise has found diverse applications in image processing, from improving the visual perception of thresholded images~\cite{PhysRevLett.78.1186} to aiding in the extraction of weak lines in noisy backgrounds using the Radon transform~\cite{ye2003sr}. Histace et al. highlighted the constructive role of Gaussian noise injection in image restoration, particularly in removing impulsive noise~\cite{histace2006constructive}. In image segmentation, adding noise has proven effective in mitigating internal noise and enhancing object detection~\cite{6487939}. Moreover, Gaussian noise addition to JPEG images not only prevents post-compression detection failures~\cite{5414609} but also assists in biometric identification by facilitating feature extraction in low-quality fingerprint images~\cite{RYU2011107}. In image transmission, controlled noise levels can improve the accuracy of coherent imaging systems~\cite{blanchard2007noise} and enable earlier detection of micro-calcifications in mammograms, potentially hastening breast cancer diagnosis~\cite{peng2009noise}. Further applications of noise enhanced image processing extend to edge detection~\cite{hongler2003resonant}, image enhancement, and object detection~\cite{ye2004image,RALLABANDI2008316,rallabandi2008enhancement, zheng2016object}, underscoring noise's multifaceted utility in advancing image processing technologies. 


%

\section{Noise in Machine Learning and Machine Learning Applications}

\noindent In the previous section, we considered the introduction of noise prior to signal/image processing for system performance enhancement. In a broader sense, we introduce randomness (adding noise was one means to introduce randomness) to achieve system performance enhancement. Our discussion in the rest of the paper will include this broader notion of introducing noise. This section describes the role of noise in performance enhancement across unsupervised and supervised learning settings, as well as in (supervised) deep learning frameworks.

\subsection{Noise-Enhanced Unsupervised Learning} 

\noindent This section describes the beneficial role of noise in enhancing the convergence rates of models/algorithms such as Markov chains and Expectation Maximization. These advantages extend to applications like clustering, Hidden Markov Models (HMMs), and neural network training, where noise facilitates faster and potentially more accurate learning outcomes
~\cite{meyn2012markov}.

\subsubsection{Markov Chain.} Convergence in Markov chains refers to the process by which the chain reaches a stable, stationary distribution over time, regardless of its starting state. The introduction of noise can expedite the convergence of discrete finite-state Markov chains towards their equilibrium states~\cite{franzke2011noise}. The following "Noise Benefit Theorem" for Markov chains illustrates how noise can lead a chain to explore new regions within the state space, thereby accelerating the convergence to the steady-state distribution.

\begin{thm}[Markov Chain Noise Benefit Theorem~\cite{franzke2011noise}]
Consider a finite, time-homogeneous Markov chain $M$ consisting of $N$ states with transition matrix $P$. Assume $M$ is irreducible and aperiodic. Let $x$ denote the nonstationary state density vector. Then, for all nonstationary state density vectors $x$, there exists a benefit in adding noise where there exists some positive $A$ such that for all $a \in (0, A)$,
$$| [\widetilde{x} P - x^\infty]_i | < |[ xP - x^\infty ]_i |$$
for every state $i$ with $\Delta_i = ( x - x^\infty ) P_i > 0$. Here, 
$$\widetilde{x} = \frac{1}{1+a}(x+n)$$ 
represents the normalized state vector post the addition of a noise vector $n$ characterized by a single nonzero element 
$$ n_j =
\begin{cases}
a & j = k \\
0 & j \neq k 
\end{cases} $$ 
for any $k$ satisfying $\Delta_k = ( x - x^\infty ) P_k > 0$.

\end{thm}
A subsequent corollary in ~\cite{franzke2011noise} further shows that a noise benefit is also observable when system states comply with an alternate norm inequality ($\Delta_i < 0$), showing the vast benefits of adding noise in improving the convergence performance Markov chain.

\subsubsection{Expectation Maximization.} The Expectation-Maximization (EM) algorithm, an iterative method for maximum-likelihood estimation from incomplete or corrupted data, benefits significantly from the introduction of noise to hasten its convergence~\cite{Dempster77maximumlikelihood, mclachlan2007algorithm}. This algorithm finds broad application in clustering~\cite{celeux1992classification}, medical imaging~\cite{shepp1982maximum}, and speech recognition~\cite{rabiner1989tutorial}. Osoba et al. introduced a noise-enhanced version of the EM algorithm, demonstrating its efficacy in accelerating convergence~\cite{osoba2013noisy}.

The EM algorithm seeks the maximum-likelihood estimate $\hat{\theta}$ of parameter $\theta$ for a given parametric probability density function (pdf) $f(y | \theta)$, based on data $Y$. The objective is
$$\hat{\theta} =  \arg \max_{\theta} \ell(\theta| y),$$
where $\ell(\theta | y) = \ln f(y | \theta)$ denotes the log-likelihood. The challenge of handling incomplete data is addressed by introducing a hidden variable $Z$, treating it as lost or unobserved data, and optimizing the log likelihood $\ell(\theta | y, z)$ to form a surrogate function $Q(\theta | \theta_k)$. Thus, the EM algorithm iteratively applies two steps to the data vector $y$:
 \begin{eqnarray*}
   \small \texttt{Expectation-Step:} &  Q(\theta |  \theta_k) & =  \mathbb{E}_Z[\ell(\theta | y,Z)| Y = y, \theta_k]\\
   \small \texttt{Maximization-Step:} & \theta_{k+1} & =  \arg \max_{\theta}  {Q (\theta | \theta_k)}
\end{eqnarray*}

Here, $Q(\theta |  \theta_k)$ represents the expectation of $\ell(\theta | y, z)$ across all possible values of $Z$ given the observation $Y = y$ and the current parameter estimate $\theta_k$. Introducing noise into the EM algorithm modifies the surrogate log-likelihood function and its maximizer to:
\begin{eqnarray*}
Q_N (\theta | \theta_k)  &= & \mathbb{E}_{Z|y,\theta_k} [\ln f(y + N,Z | \theta)],\\
\theta_{k+1,N}  &=  & \arg \max_{\theta}  {Q_N (\theta | \theta_k)}.
\end{eqnarray*}

Assuming $Q(\theta | \theta_*)$ as the final surrogate log-likelihood given the optimal EM estimate $\theta_*$, where $\theta_*$ maximizes $Q(\theta | \theta_*)$, and $Q( \theta_* | \theta_*) \geq Q(\theta | \theta_*) $ for all $\theta$, a noise benefit is observed when:
$$Q_N (\theta_k | \theta_*) \geq Q(\theta_k | \theta_*),$$
or equivalently,
$$(Q(\theta_* | \theta_*) - Q(\theta_k |\theta_*)) \geq (Q(\theta_* |\theta_*) - Q_N( \theta_k | \theta_*)).$$

The following theorem outlines the conditions under which noise benefits the EM algorithm.

\begin{thm}~\label{theo:nem}(Expectation-Maximization Noise Benefit Theorem (NEM Theorem)~\cite{osoba2013noisy}). Consider a noise random variable $N$ with a pdf $f(n | y)$, and a sequence of EM estimates $\{\theta_k\}$ for $\theta$, with $\theta_* = \lim_{k \to +\infty} \theta_k$ being the converged EM estimate. Assume the injected noise keeps the data within the support of the likelihood function. A noise benefit in an EM estimation iteration,
$$(Q(\theta_* | \theta_*) - Q(\theta_k | \theta_*)) \geq (Q(\theta_* | \theta_*) - Q_N (\theta_k | \theta_*)),$$
occurs on average if the following expectation holds:
\begin{equation}
\mathbb{E}_{Y,Z,N|\theta_*} \Big[ \ln{\Big(\frac{f(Y + N,Z | \theta_k)}{f(Y,Z | \theta_k)}\Big)}\Big] \geq 0.\label{eq:expectationTerm}
\end{equation}
\end{thm}
This expectation term guarantees that the noise-perturbed EM, on average, yields larger improvements on estimates than the noiseless EM after any $k$ steps~\cite{osoba2013noisy}. As we detail next, more specific conditions can be provided under which noise is beneficial in mixture density models and log-convex probability density models (see further details in~\cite{osoba2013noisy}).

Given the prevalence of mixture models in EM applications, we highlight the form of noise that can be beneficial in mixture models based on NEM theorem~\cite{osoba2013noisy}. Clearly, having a dominated-density condition ensures Equation (\ref{eq:expectationTerm}) is positive. That is, 
\begin{equation*}
\mathbb{E}_{Y,Z,N|\theta_*} \Big[ \ln{\Big(\frac{f(Y + N,Z | \theta_k)}{f(Y,Z | \theta_k)}\Big)}\Big] \geq 0,
\end{equation*}
when $$f(y + n, z | \theta)\geq f(y, z | \theta)$$ for nearly all $y$, $z$, and $n$. Based on this understanding, we can identify the form of noise $N$ that yields NEM benefits to mixture models.

\begin{cor}[NEM Condition for Gaussian Mixture Models (NEM-GMM)~\cite{osoba2013noisy}]
For Gaussian mixture model data where $Y|_{Z=j} \sim \mathcal{N}(\mu_j, \sigma_j^2)$, described by the normal pdf $f(y | j, \theta)$, the following condition
$$f(y + n | j, \theta) - f(y | j, \theta) = \Delta f_j(y, n)\geq 0 $$
is met if
$$n^2 \leq 2n(\mu_j - y).$$
\end{cor}

Hence, under the Gaussian Mixture Model–NEM (GMM–NEM) condition, the NEM algorithm estimates the standard deviations $\sigma_j$ faster than the traditional EM approach. The GMM–NEM condition dictates that noise $n$ should fall within one of two possible intervals:
\begin{eqnarray*}
N_j^- (y) &=& [2(\mu_j - y), 0],\\
N_j + (y) &=& [0, 2(\mu_j - y)],  
\end{eqnarray*}
suggesting that noise $N$ tends to shift the data sample $y$ away from distribution tails towards the mean clusters. Osoba et al.~\cite{osoba2013noisy} demonstrated that larger sample sizes decrease the probability of satisfying the NEM theorem within Gaussian Mixture Models. Intriguingly, noise demonstrates enhanced efficacy in sparse data settings, effectively generating pseudo-data. This principle of noise enhancement in the EM algorithm (NEM algorithm) leads to broad applications in various domains, as illustrated in the following examples.\vspace{3mm}

\noindent{$\blacktriangleright$ \textit{I. Noise Benefit for Clustering.} Given $k$ clusters in the dataset, with $\theta_1, \dots, \theta_k$ representing the pdf parameters for each cluster, and $\alpha_1, \dots, \alpha_k$ as the mixing proportions, EM clustering employs the membership probability density function $p_Z (j|y, \Theta_{EM})$ as a  maximum a posteriori classifier for each sample $y$. Here, the label is the cluster label. A sample $y$ is assigned to the $j$-th cluster if $p_Z (j|y, \Theta_{EM})\geq p_Z (k|y, \Theta_{EM})$, $\forall k \neq j$; hence,
$$\text{EM-class}(y) = \arg \max_{j} p_Z (j|y, \Theta_{EM}),$$
where $$p_Z (j|y, \Theta) = \frac {\alpha_j  f(y|Z = j, \theta_j)} {f(y|\Theta)},$$ for $\Theta = \{\alpha_1, \dots , \alpha_K , \theta_1, \dots , \theta_K\}$. Noise-enhanced EM clustering follows the same classification logic but uses NEM-optimal GMM parameters:
$$\text{NEM-class}(y) = \arg \max_{j} p_Z (j|y, \Theta_{NEM}).$$
Osoba et al. provided evidence that the GMM Expectation–Maximization algorithm encompasses $k$-Means clustering~\cite{osoba2013noise}. Thus, the benefits of noise in the NEM Theorem (Theorem ~\ref{theo:nem}) extend to EM-clustering. The next theorem elucidates the clustering noise benefits in terms of reduced ``misclassification" relative to the EM-optimal classifier, where misclassification here is defined as the normalized number of disagreements. Let $\text{class}_{opt} (Y) = \arg \max p_Z (j|Y, \theta_*)$ denote the EM-optimal classifier using the optimal model parameters $\theta_*$. Define $P_M [k] = P (\text{EM-class}_k(Y) \neq \text{class}_{opt}(Y))$ and $P_{M_N} [k] = P (\text{NEM-class}_k(Y) \neq \text{class}_{opt}(Y))$ as the probabilities of misclassification in EM and NEM clustering, respectively, relative to $\text{class}_{opt}$, using the $k$-th iteration parameter.

\begin{thm}[Clustering Noise Benefit Theorem~\cite{osoba2013noise}]
At the $k$th iteration, the NEM misclassification probability $P_{M_N}[k]$ is less than or equal to the EM misclassification probability $P_M[k]$:
$$P_{M_N}[k] \leq P_M[k],$$
provided the additive noise $N$ in the NEM-clustering process satisfies the NEM Theorem condition:
$$\mathbb{E}_{Y,Z,N|\theta_*} \Big[ \ln{\Big(\frac{f(Y + N,Z | \theta_k)}{f(Y,Z | \theta_k)}\Big)}\Big] \geq 0.$$
\end{thm}
This theorem leads to a simple algebraic condition for noise benefit in each coordinate $i$:
$$n_i[ n_i - 2n_i (\mu_{j_i} - y_i) ] \leq 0 $$ for all $j$. Simulations show that noise accelerates convergence in various types of competitive learning—supervised, unsupervised, and differential—under broad general conditions. Here, the independent noise is modeled as a sequence of random white Gaussian vectors $n$ with diminishing variance. Osoba et al. later generalized this theorem to encompass arbitrary measurable noise injections $\phi(Y,N)$, provided the noise continues to meet the NEM positivity condition~\cite{osoba2016noisy}.\vspace{3mm}

\noindent{$\blacktriangleright$ \textit{II. Noise Benefit for Hidden Markov Models (HMMs).} Hidden Markov Models (HMMs) represent another domain where the introduction of noise can yield significant benefits~\cite{audhkhasi2013noisy}. HMMs are probabilistic models often used for sequencing time series data that have found widespread application in fields such as speech processing and recognition~\cite{rabiner1989tutorial,wilpon1990automatic}. The integration of noise is advantageous for HMMs since the \textit{Baum-Welch algorithm}—a variant of the EM algorithm used for training HMM parameters through maximum likelihood estimation—can harness the enhancements offered by the NEM theorem.\vspace{3mm}

\noindent{$\blacktriangleright$ \textit{III. Noise Benefit for Neural Networks.}
Similarly, Neural Networks can benefit from the principles of NEM. Audhkhasi et al. have demonstrated that the backpropagation algorithm can be considered a specialized form of the generalized EM algorithm~\cite{audhkhasi2013noise}. The application of noise to both hidden and output neurons, as later explored in Sections~\ref{section:NNHiddenNeuron} and~\ref{section:NNOutputNeuron}, further highlights the broad applicability of NEM principles within neural network training processes. 
As we transition from the noise-enhanced unsupervised learning part, where randomness helps in uncovering hidden patterns without labels, we now enter the structured world of noise-enhanced supervised learning. Here, noise continues its role as a facilitator, this time refining models trained on labeled data to achieve greater generalization and robustness.

\subsection{Noise in Deep Learning} 
\label{section:DL}
This section explores the significant impact of adding noise during the training process on enhancing the performance of neural networks for classification and regression tasks. Introducing random noise to inputs, weights, gradients, or hidden units is a long-standing practice among neural network practitioners~\cite{an1996effects}. We provide an overview of notable contributions below.

\subsubsection{Adding Noise to Input.}~\label{section: NNInputNeuron} Regularization is a common strategy to improve a network's generalization ability and to manage the model's \textit{bias-variance} tradeoff~\cite{shalev2014understanding}. Bishop et al. have shown that injecting random noise into input data serves as an alternative form of regularization, akin to \textit{Tikhonov regularization}, enhancing the network's ability to generalize~\cite{bishop1995training}. Similarly, introducing noise into the training data of feedforward neural networks can improve weight vector estimation and model generalization~\cite{holmstrom1992using}, essentially simulating a kernel estimate of the probability density of the training vector distribution.

\textit{SMOOTHGRAD} is a technique that adds noise to images during training to diminish visual noise, aiming to highlight pixels significantly influencing classification decisions~\cite{smilkov2017smoothgrad}. The method averages the sensitivity maps generated from slight perturbations of an input image, thereby smoothing the gradients and producing a denoising effect from noise addition during training.

\textit{Denoising Autoencoders} benefit from noise injection by learning robust representations through the reconstruction of clean inputs from their corrupted versions~\cite{vincent2010stacked}. The corruption usually involves adding Gaussian or masking noise to the input, forcing the autoencoder to distill essential data features into more useful representations.\vspace{1mm}

\noindent{$\blacktriangleright$ \textit{Data Augmentation.}
Noise addition to neural network inputs can also be seen as \textit{data augmentation}~\cite{goodfellow2016deep}, particularly useful when dealing with small datasets. By creating a noisy dataset from the original one, this augmentation technique helps address the challenge of limited data for certain classes. For instance, the study~\cite{akbiyik2023data} evaluates the impact of various noise models on images for different Convolutional Neural Network architectures, emphasizing data augmentation's utility in image classification. Moreover, data augmentation proves beneficial for non-image data, albeit with a lesser effect on enlarging training set size. Moreno-Barea et al. propose a forward noise adjustment scheme to enhance prediction accuracy in non-image data classification, showing a notable increase in accuracy across several UCI benchmark datasets~\cite{moreno2018forward}. Section~\ref{section:graphAugmentation} provides more details about augmentation methods in the graph domain. 

Generative Adversarial Networks (GANs) represent another intriguing application where noise serves as the sole input, generating diverse data through a generator-discriminator dynamic~\cite{goodfellow2014generative}. While input noise injection is prevalent, noise can also effectively be applied to other network components during training, such as activation layers, weights, gradients, and outputs, broadening the scope of noise-enhanced learning strategies.

\subsubsection{Adding Noise to Output Neurons.}~\label{section:NNOutputNeuron} This section discusses how NEM noise injection into output neurons can facilitate faster convergence in networks designed for classification or regression tasks~\cite{audhkhasi2013noise}.\vspace{1mm}

\noindent\textit{Classification.} We examine the noise benefit for feedforward neural networks utilizing Gibbs or softmax activation in output neurons for $K$-class classification. Let $x$ represent input layer neuron values, and $h$ denote hidden layer neuron values. For a $K$-valued target variable $y$, with $1$-in-$K$ encoding denoted as $t$, where $t_k = 1$ for the correct class and $0$ otherwise, we introduce noise $n$ to $t$ to establish the hyperplane noise condition beneficial for networks that perform classification:

\begin{thm}[Hyperplane Noise Benefit Condition for Feedforward Neural Networks~\cite{audhkhasi2013noise}]~\label{thm:hyper} The NEM Theorem condition for maximum likelihood training of feedforward neural networks with Gibbs or softmax activation output neurons is met if
$$\mathbb{E}_{t,h,n|x,\Theta_*} \Big\{ n^T \log (a^t) \Big\} \geq 0,$$
where $\log (a^t)$ indicates the log-activation vector of the output neurons.
\end{thm}

In essence, for beneficial noise impact, $n$ should lie above a hyperplane defined by normal $\log(a^t)$, accelerating the training process.\vspace{1mm}

\noindent\textit{Regression.} For networks performing regression that use Gaussian output neurons, there exists a spherical region in the noise space where noise injection yields a positive effect:

\begin{thm}[Sphere Noise Benefit Condition for Feedforward Neural Networks~\cite{audhkhasi2013noise}] The condition for NEM theorem applicability in maximum likelihood training of feedforward networks with Gaussian output neurons is
$$\mathbb{E}_{t,h,n|x,\Theta_*} \Big\{ \left\|n - a^t + t\right\|^2 - \left\|a^t - t\right\|^2 \Big\} \leq 0,$$
where $\left\| . \right\|$ denotes the $L2$ vector norm.
\end{thm}

Thus, noise $n$ should reside within a sphere centered at $t - a^t$ with radius $\left\|t - a^t\right\|$ to expedite backpropagation convergence.\vspace{2mm}

Noise benefits extend to Convolutional Neural Networks (CNNs)~\cite{audhkhasi2016noise}. For CNNs with a hidden convolutional layer, let $X$ represent the input $2$-dimensional data, and $y$ the $1$-in-$K$ binary encoding vector of the target label. Denote $Z_j$ as the sigmoid activation of the $j$th hidden neuron and $U_j^k$ as the weight matrix connecting hidden neurons to output neurons. The following theorem describes the sufficient condition for noise benefits in CNNs by introducing NEM noise solely to the $1$-in-$K$ encoding vector $y$ of target class labels:

\begin{thm}[Hyperplane Noise-Benefit Condition for CNNs~\cite{audhkhasi2016noise}).] Maximum likelihood training of a CNN with softmax or Gibbs activation output neurons satisfies the NEM Theorem condition if
$$\mathbb{E}_{y,Z_1, ... , Z_j,n|X,\Theta_*} \Big\{ n^T \log (a^t) \Big\} \geq 0,$$
where $a_k^t$ is the activation of the $k$th output neuron, defined as
$$a_k^t = \frac{\exp\Big({\sum^{J}_{j=1} e^T Z_j \odot U_j^k e }\Big)}{\sum^{K}_{k_1=1} \exp\Big({\sum^{J}_{j=1} e^T Z_j \odot U_j^{k_1} e }\Big)},$$
with $\odot$ representing the element-wise Hadamard product, and $e$ is a vector of all $1$s. 
\end{thm}

Noise above the hyperplane defined by $\Big\{ n^T \log (a^t)=0 \Big\}$ can, on average, accelerate CNN training. Injecting noise into the output neurons decreases the average per-iteration training-set cross-entropy and classification error.

\subsubsection{Adding Noise to Hidden Neurons}~\label{section:NNHiddenNeuron}
The concept of NEM noise-benefit extends to the injection of noise into hidden neurons as well~\cite{kosko2020noise}. Given that hidden neuron activations become observable data during a network's forward pass, each hidden layer effectively acts as an output layer for preceding hidden layers. There are two primary methods for injecting NEM noise into hidden neurons~\cite{kosko2020noise}: (1) transferring the same NEM noise used in the output layer to a hidden layer, and (2) introducing distinct NEM noise to the hidden layer, independent of the output layer. We explore the first approach before delving into the theorem governing the general case.

\begin{itemize}
    \item[a] \textit{Injecting the Same NEM Noise from the Output Layer~\cite{kosko2020noise}:} Suppose NEM noise $n$ added to the output layer is also to be introduced into the final hidden layer $h_k$. Injecting NEM noise $n$ into the output targets $t$ effectively adds $n$ to the error $e^t$:
$$e^t = t - a^t , e^t_N =  t + n - a^t= e^t + n.$$ 
This noisy error vector $e^t_N$ then backpropagates over the weights to the hidden layer, treating hidden neuron activations as visible data. This transformation of noise satisfies the NEM sufficient condition in logistic hidden layers if
$$(\mathbf{U}n)^T \ln a^h \geq (\mathbf{U}n)^T \ln(1 - a^h),$$
where $a^h$ represents the activation vector of the hidden layer and $\mathbf{U}$ the weight matrix between the $J$ hidden neurons and the $K$ output neurons.
\item[b] \textit{Introducing New NEM Noise to Hidden Neurons:} The theorem below articulates the noise benefit when introducing new NEM noise to logistic hidden neurons:
\begin{thm}[Noise-benefit in Hidden Logistic Neurons~\cite{kosko2020noise})]. NEM noise $n$ benefits a layer of logistic neurons if the injected noise fulfills the NEM likelihood inequality
$$\mathbb{E}_{h,n|x,\Theta_*} \Big\{ n^T \ln a^h \Big\} \geq \mathbb{E}_{h,n|x,\Theta_*} \Big\{ n^T \ln (1-a^h) \Big\},$$
with $a^h$ denoting the logistic layer's hidden-layer activation vector.
\end{thm}

Incorporating noise into hidden neuron activations during training can make the network more robust to disruptions or errors that might occur in the internal components, such as unreliable or faulty nodes~\cite{murray1994enhanced}. Similarly, randomly disabling hidden nodes during training can mitigate the impact of node failures in multilayer perceptrons~\cite{sequin1990fault}. Poole et al. expanded the use of noise in autoencoders, applying it not only at the input layer but throughout the network, especially prior to non-linear activations and within hidden units. This approach fosters a comprehensive framework for various regularization strategies, extending beyond conventional denoising techniques~\cite{poole2014analyzing}.
\end{itemize}

\subsubsection{Adding Noise to Network Weights.} Supervised neural networks often generalize more effectively when the information content in the network weights is significantly less than that in the training output vectors. One approach to managing the informational content of weights involves the addition of Gaussian noise~\cite{168304.168306}. This technique simplifies neural networks by diminishing the informational requirements for transmitting parameters, thereby enhancing model generalization. Murray and Edwards~\cite{317730} demonstrated the advantages of simulating synaptic noise during the training of multilayer perceptrons by introducing stochastic variations into the synaptic weights at each layer. Their findings suggest that such noise can improve fault tolerance, generalization performance, and the learning trajectory of the network. Kam-chuen et al. explored the benefits of synaptic noise in the weight training of recurrent neural networks, observing enhancements in generalization and convergence performance, and in some cases, both~\cite{jim1996analysis}. Specifically in Deep RNNs used for speech recognition, the injection of Gaussian noise into LSTM weights for each training sequence has been shown to reduce generalization errors~\cite{graves2013speech}, acting as a regularization mechanism to prevent overfitting.

\subsubsection{Adding Noise to Network Gradients}
While the practice of adding random noise to classical neural networks is longstanding, the specific benefits of gradient noise in modern deep networks have been less explored. Neelakantan et al. introduced a novel technique involving the injection of annealed Gaussian noise into the training gradients to enhance the training and generalization of complex neural networks~\cite{neelakantan2015adding}. This method, which aims to mitigate overfitting and reduce training loss, encourages a more active exploration of the parameter space, distinguishing itself from weight noise by the annealing of the noise to zero at convergence.

The critical function of noise in stochastic gradient descent for selecting optimal solutions~\cite{Bottou91stochasticgradient}, along with the potential of noisy gradients to augment deep neural network training~\cite{neelakantan2015adding}, prompted further theoretical examination into noise's role in DNN training. Perturbed gradient descent with noise annealing has been theoretically validated to escape from suboptimal local minima and converge to global optima within polynomial time, irrespective of the initialization~\cite{zhou2019toward}. 


\subsection{Noise in Applied Machine Learning}
This section explores the beneficial roles of noise in enhancing machine learning applications, particularly in machine learning with graphs, natural language processing, and recommender systems. Examples include how noise can improve graph connectivity, assist community detection and link prediction, and make language models more robust to real-world variations.

\subsubsection{Graph Machine Learning.} 

In the rapidly progressing area of machine learning with graphs and network science, noise can play a significant role, from reshaping graph structures to enhancing connectivity and community detection. Adding noise in graph machine learning refers to modifying the structure or attributes of graphs by randomly adding, removing, or altering edges, nodes, or attributes.\vspace{4mm}

\noindent{$\blacktriangleright$ \textbf{I. Noise-Enhanced Graph Connectivity.}} Enhancing graph connectivity by adding edges has a long history, aiming at increasing the \textit{algebraic connectivity} of graphs. Ghosh et al. investigated strategies for edge addition to maximize a graph's algebraic connectivity, defined as the second smallest eigenvalue $\lambda_2(L)$ of the Laplacian matrix of a graph~\cite{ghosh2006growing}. Given a base graph $G_{base} = (V,E_{base})$, a set of candidate edges $E_{\text{candidate}}$ among nodes in $V$, and a target number of edges $k$ ($0 \leq k \leq m_c$) to be added ($k$ noisy edges) for optimal connectivity enhancement, the problem is framed as:

\begin{equation*}
\begin{array}{ll@{}ll}
\text{maximize}  & \lambda_2~(L(E_{base} \cup E)) \\
\text{subject to}& |E| = k,\quad E \subseteq E_{\text{candidate}}  
\end{array}
\end{equation*}

Ghosh et al. proposed a greedy heuristic approach to tackle this problem by: (1) identifying a unit eigenvector $v$ associated with $\lambda_2(L)$ (where $L$ is the current Laplacian matrix) and (2) adding an edge $e \sim (i, j)$ that maximizes the value of $(v_i - v_j)^2$ to the graph. This procedure repeats until $k$ edges are added, effectively (1) linking separate connected components, (2) connecting nodes strongly affiliated with different components (where $(v_i - v_j)^2$ is large), and (3) joining nodes most distant from each other in the linear embedding. In graph theory, linear embedding refers to the process of mapping graph nodes onto a one-dimensional space by assigning each node a coordinate such that connected nodes are positioned close together.


\begin{figure}[t]
\centering
\subfloat[original graph]{\label{fig:A}
\resizebox{6cm}{!}{
\begin{tikzpicture}
\draw[color=blue, fill=blue, very thick] (-1.35,2.05) ellipse (0.75 and 0.55);
\draw[color=white, fill=white, very thick] (-1.35,2.05) ellipse (0.7 and 0.5);
\draw[color=blue, fill=blue, very thick] (0.45,2) ellipse (0.75 and 0.5);
\draw[color=white, fill=white, very thick] (0.45,2) ellipse (0.7 and 0.45);
\draw[color=blue, fill=blue, very thick] (1.98,1.92) ellipse (0.55 and 0.85);
\draw[color=white, fill=white, very thick] (1.98,1.92) ellipse (0.5 and 0.8);

\fill[black] (-1.7,2.5) circle (0.05cm) node[above]{$1$};
\foreach [count=\index from 2] \x in {-2,-1.4,-0.8,-0.2,0.4,1}
    \fill[black] (\x, 2) circle (0.05cm) node[below left] {\index};
\fill[black] (1.6,2) circle (0.05cm) node[below]{$8$};
\fill[black] (2.2,2.5) circle (0.05cm) node[below]{$9$};
\fill[black] (2.2,1.5) circle (0.05cm) node[below left]{$10$};

\draw[black,thick] (-2,2) -- (1.6,2);
\foreach \x in {-2, -1.4}
    \draw[black, thick] (-1.7, 2.5) -- (\x,2);
\foreach \y in {1.5,2.5}
    \draw[black,thick] (1.6,2) -- (2.2,\y);
\end{tikzpicture}
}
}%
\subfloat[noisy graph]{\label{fig:B}
\resizebox{6cm}{!}{
\begin{tikzpicture}
\draw[color=green, fill=green, very thick] (3.2,2.07) ellipse (1.05 and 0.56);
\draw[color=white, fill=white, very thick] (3.2,2.07) ellipse (1.02 and 0.53);
\draw[color=green, fill=green, very thick] (5.78,2) ellipse (1.2 and 0.8);
\draw[color=white, fill=white, very thick] (5.78,2) ellipse (1.15 and 0.75);

\fill[black] (2.6,2.5) circle (0.05cm) node[above]{$1$};
\foreach [count=\index from 2] \x in {2.3,2.9,3.5,4.1,4.7,5.3, 5.9}
    \fill[black] (\x,2) circle (0.05cm) node[below left]{\index};
\fill[black] (6.5,2.5) circle (0.05cm) node[below]{$9$};
\fill[black] (6.5,1.5) circle (0.05cm) node[above]{$10$};
\draw[black,thick] (2.3, 2) -- (5.9, 2);
\foreach \x in {2.3,2.9}
    \draw[black,thick] (2.6, 2.5) -- (\x,2);
\foreach \y in {2.5,1.5}
    \draw[black,thick] (5.9, 2) -- (6.5,\y);
\draw[blue, dash dot,thick] (2.9, 2) .. controls(3.5,2.35) .. (4.1, 2);
\end{tikzpicture}
}
}%
\hfill

\caption{Detected Communities (ovals) before (a) and after (b) adding noise (dashed edge) using the same community detection algorithm (Leading Eigenvector method). Adding noise edge $(3,5)$ in (b) helps find better communities, as observed by $30\%$ decrease in the value of community detection objective function (edge cut, here).}
\label{fig:noiseExample}
\end{figure}
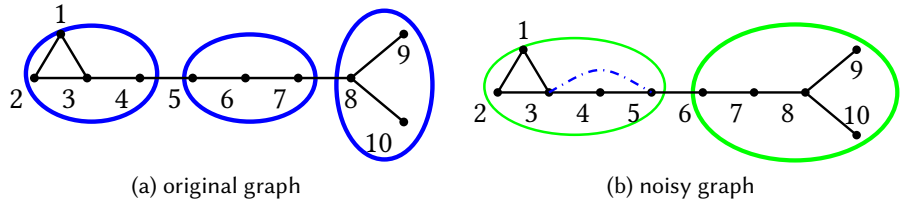


\vspace{2mm}
\noindent{$\blacktriangleright$ \textbf{III. Noise-Enhanced Community Detection and Link Prediction.} Abdolazimi et al. have explored enhancing graph algorithms through strategic noise addition to graphs. They proposed a framework that introduces a controlled level of randomness into community detection and link prediction algorithms by integrating noisy edges into the network~\cite{abdolazimi2020noise, abdolazimi2021noise}. The primary aim of this framework is to improve the efficiency of community detection and the accuracy of link prediction without significantly increasing computational demands.

To demonstrate how noise addition can enhance community detection, an illustrative example is presented. Consider the graph in Figure \ref{fig:A} with $10$ nodes and $10$ edges. Using the Leading Eigenvector community detection method~\cite{newman2006finding}, three communities are identified: $\{1, 2, 3, 4\}$, $\{5, 6, 7\}$, and $\{8, 9, 10\}$. These communities are evaluated using the \textit{edge cut} objective function~\cite{leskovec2010empirical}, which yields an edge cut value of 1.3. By adding a single noise edge $(3, 5)$ (illustrated as a dashed line), the \textit{noisy graph} in Figure \ref{fig:B} now shows two communities: $\{1, 2, 3, 4, 5\}$ and $\{6, 7, 8, 9, 10\}$. This modification not only reduces the number of communities but enhances their quality, as evidenced by the reduced edge cut value to 1.0, representing a 30\% improvement.

Their work introduces three main methods to incorporate noise into graphs, especially targeting nodes with high degrees:

\begin{enumerate}
    \item[] \textit{I. Random Noise:} Edge endpoints are randomly selected from available candidates.
    \item[] \textit{II. Weighted Noise:} Each node for edge endpoints is chosen based on the proportion of its degree relative to other nodes. This method weights the probability of selecting a node by its degree, enhancing the likelihood that higher-degree nodes are chosen as endpoints.
    \item[] \textit{III. Frequency Noise:} Nodes are selected based on the degree distribution among the candidates, where nodes with more common degrees are less likely to be selected. This approach aims to diversify the connectivity by incorporating less commonly connected nodes, thereby potentially introducing novel structural features into the graph.
\end{enumerate}

\vspace{2mm}
\noindent{$\blacktriangleright$ \textbf{IV. Graph Data Augmentation.}~\label{section:graphAugmentation}
As discussed in the Section~\ref{section: NNInputNeuron}, adding noise to neural network
inputs can be viewed as data augmentation~\cite{goodfellow2016deep}. Data augmentation plays a pivotal role in the realm of complex graph data, serving as a critical tool for enriching the training dataset through modifications of existing samples or the creation of synthetic ones. This process is akin to noise enhancement, as both strategies aim to introduce modified (or more varied) training examples, thereby facilitating improved model learning and generalization. Data augmentation achieves this data modification by integrating slight alterations or entirely novel examples into the dataset, helping prevent model overfitting and encouraging the recognition of more generalized patterns for better performance on unseen data.

Data augmentation techniques for graphs are categorized into three primary groups based on the type of augmented information: ($\bigstar$) \textit{structure-wise}, ($\blacksquare$) \textit{feature-wise}, and ($\spadesuit$) \textit{label-wise}. These methods respectively focus on altering the graph structure through the addition or removal of nodes/edges, perturbing or generating new node features, and enriching sparse labeled data to bolster model robustness and performance.

($\bigstar$) Structure-wise augmentation techniques inject variability by changing the connectivity patterns within graphs. A well-known example is \textit{DropEdge}~\cite{Rong2020DropEdge}, which randomly drops edges.  Another example is \textit{AdaEdge}~\cite{Chen_Lin_Li_Li_Zhou_Sun_2020}, which dynamically adjusts the graph's topology through selective edge modifications based on feedback from a graph neural network classifier; hence, targeting the \textit{over-smoothing} often observed in graph neural networks. The \textit{GAUG} framework~\cite{zhao2021data} uses neural edge predictors to predict new edges in the graph. This approach adds connections between nodes of the same class and reduce connections between nodes of different classes. This helps create groups in the graph where nodes of the same class are more closely connected.

($\blacksquare$) In feature-wise augmentations, methods such as SR+DR~\cite{song2021topological} integrate unsupervised learning-derived topological embeddings with existing node features through a dual graph neural network, enriching node representations for enhancing graph analysis. \textit{Local Augmentation}~\cite{liu2022local} generates new node features from a distribution formed based on the node's features, diversifying the input to graph neural networks in each training cycle.

($\spadesuit$) Label-wise augmentation methods, such as \textit{Label Mixing}, utilize the Mixup technique~\cite{zhang2017mixup} to generate additional training samples by linearly interpolating between pairs of training examples. This approach not only expands the training set but also aligns linear combinations of features with corresponding label combinations, enhancing the training variety.

These augmentation strategies reveal the strategic role that noise can play in graph machine learning, expanding training data to provide models with a deeper understanding of graph structure and dynamics. These augmentation methods can be used across various GNN architectures, to highlight their significance and their broad applicability.

\subsubsection{Natural Language Processing (NLP)}~\label{section:NLP}

The concept of noise injection as a form of data augmentation, previously explored in the context of neural networks (Section~\ref{section: NNInputNeuron}), extends its utility into the domain of Natural Language Processing (NLP)~\cite{goodfellow2016deep}. While data augmentation has seen widespread application in fields such as speech data analysis and computer vision, the augmentation of text data remains a more challenging endeavor due to the complexity of generating realistic textual content~\cite{Coulombe2018TextDA}. Coulombe et al. investigate the adaptation of successful data augmentation techniques from computer vision to NLP, particularly those involving data transformations in the pre-processing stage~\cite{Coulombe2018TextDA}. \textit{Textual noise injection}, a method that introduces alterations in texts---such as adding, deleting, or modifying letters within words, and changing punctuation---stands out as a prominent technique for augmenting text data. The specifics of noise injection vary, primarily in terms of the method and location of application~\cite{xie2020unsupervised,Coulombe2018TextDA, abs-1901-11196}.

For instance, \textit{Spelling Errors Injection} creates texts filled with common misspellings, thereby conditioning models to be more tolerant of such errors~\cite{Coulombe2018TextDA}. \textit{Unigram Noising}, another technique, involves replacing words with alternatives sampled from a unigram frequency distribution~\cite{xie2020unsupervised}. \textit{Blank Noising} substitutes a random word in a sentence with a placeholder token~\cite{xie2020unsupervised}. Furthermore, the utility of noise is shown in automatic speech recognition, where methods developed for noise robustness have significantly improved the understanding of spoken language in noisy environments. This performance improvement highlights the importance of noise-aware processing in NLP systems that deal with spoken input, providing a bridge between audio signal processing and textual understanding~\cite{hu2024large}. A notable implementation of noise-enhanced NLP is found in the Python library \texttt{NoiseMix}, which diversifies datasets by applying variations to copies of existing data entries~\cite{noisemix}. This library systematically introduces textual noise into the data.

\noindent{$\blacktriangleright$ \textit{Noise in Large Language Models.}} 
Recent advancements in Large Language Model (LLM) optimization have demonstrated the power of controlled noise in improving model performance and robustness. Methods like NEFTune and GIFT-SW have demonstrated the power of controlled noise in fine-tuning LLMs~\cite{jain2023neftune, zhelnin2024gift}. NEFTune which is a noise-enhanced finetuning method, has demonstrated that introducing controlled noise into embedding vectors can significantly improve the generalization ability of LLMs~\cite{jain2023neftune}. For instance, when applied to models like LLaMA-2, NEFTune boosted performance in conversational settings, with accuracy improvements of up to 35\% on AlpacaEval. This approach reduces overfitting to instruction datasets by introducing variations during finetuning, encouraging the model to generate more diverse and flexible responses. GIFT-SW, on the other hand, focuses on updating only salient weights while injecting Gaussian noise into non-salient weights~\cite{zhelnin2024gift}. This noise injection helps improve convergence and model robustness, particularly during optimization and quantization processes. Experiments show that this method enhances performance, reduces overfitting, and improves stability, making it more computationally efficient than other fine-tuning approaches.

Expanding on this concept, Zheng et al.~\cite{zheng2023noisy} investigated the impact of domain-agnostic perturbations on LLMs' performance in multi-hop reasoning tasks. They found that increasing the proportion of perturbed exemplars in few-shot prompting methods generally improved the models' robustness to perturbations in test questions. This suggests that exposing LLMs to noisy examples during the few-shot learning process can enhance their ability to handle real-world variations in input, such as typos or semantic variations.

The benefits of controlled noise extend to other NLP techniques that can enhance LLM capabilities. Retrieval-Augmented Generation (RAG) models demonstrate how integrating external "noise" in the form of additional data sources can significantly improve the robustness and quality of generated outputs in knowledge-intensive tasks ~\cite{lewis2020retrieval}. By dynamically retrieving and incorporating relevant information from a vast database during the generation process, RAG shows how external data can serve as beneficial noise, improving the factual accuracy and depth of responses in question-answering and other NLP applications. Expanding on this, recent findings in RAG systems reveal another fascinating application of noise. Studies have shown that incorporating random, unrelated documents into the retrieval process paradoxically enhances model accuracy by up to 35\%. This highlights that noise can serve as a beneficial factor in preventing overfitting to overly specific prompts, thus leading to improved LLM robustness~\cite{cuconasu2024power}.

In broader NLP applications, such as neural language generation, noise injection sampling and self-training have been shown to reduce attribute realization errors in dialogue generation tasks~\cite{kedzie2019good}. This is achieved through data augmentation, where noise-enhanced meaning representation/text utterance pairs help models better capture the nuances of dialogue. Similarly, Vaibhav et al. introduce synthetic noise into clean data for machine translation. This approach makes systems more resilient to variances found in natural, noisy text such as social media content. It significantly improves translation accuracy without requiring naturally noisy parallel data~\cite{vaibhav2019improving}.

\subsubsection{Recommendation Systems}

Noise injection has also been used in industry-deployed recommender systems~\cite{resnick1997recommender}. Rather than consistently displaying the system's top-recommended content to users, recommender systems occasionally present random content with a low probability. The approach offers multiple benefits for improving recommendation systems: (1) It mitigates system bias~\cite{mansoury2020feedback}. Recommendation systems often exhibit feedback loops and biases, wherein certain content is continuously recommended, reinforcing positive feedback loops, and marginalizing exposure for other content. Consequently, the true quality of the alternative content remains unclear. Introducing randomness into the system helps alleviate bias to some extent; (2) A primary objective of recommendation systems is to evaluate and predict the value of content for ranking and presentation to users. Recent advances have employed causal inference techniques~\cite{wang2020causal} to estimate the causal effects of content. Therefore, \textit{randomized controlled trials (RCTs)}, recognized as the gold standard for causal inference, can be used. By incorporating noise (i.e., displaying random content), RCTs can be implemented to measure the difference (or uplift) between users viewing recommended content versus random content~\cite{gordon2023predictive}. Given that noise is introduced with a low probability, the cost of conducting RCTs can be effectively managed.


\section{Other Applications}
This section covers additional uses of noise beyond the conventional fields discussed thus far, extending into natural sciences, optimization, and privacy preservation.

\subsection{Noise Applications in Natural Sciences}
Noise has shown various benefits in various natural sciences, including a strong presence in quantum physics, neuroscience, and biology.

\subsubsection{Noise-Enhanced Quantum Physics.} 
Quantum physics, the study of the behavior of matter and energy at the smallest scales, faces unique challenges distinct from those in classical physics. One such challenge is quantum noise, which arises due to the fundamental uncertainty inherent in quantum systems and the interaction of these systems with their environment. Unlike classical noise, which can often be reduced or mitigated through improved measurement techniques or signal processing, quantum noise is deeply rooted in the principles of quantum mechanics and decoherence. Decoherence, a process where the quantum system loses its quantum properties due to interaction with the environment, poses a significant hurdle in the development and performance of quantum information processing systems. 

Addressing quantum noise is crucial for the advancement of quantum computing, quantum cryptography, and other quantum technologies. In this context, stochastic resonance (SR) emerges as a counterintuitive yet promising approach. SR in quantum systems leverages noise, enhancing the information processing capabilities of quantum systems rather than detracting from them. In this section, we explore the beneficial effects of SR in quantum physics, highlighting key research findings and applications that demonstrate the potential of noise enhancement in overcoming the challenges posed by quantum noise.

Various forms of stochastic resonance have been reported for information communication over qubit channels in the presence of noise~\cite{ting1999stochastic, bowen2004noise, bowen2006stochastic}. For instance, Bowen et al. have demonstrated that increasing noise levels within quantum channels can, paradoxically, augment the rate of information transmission in both quantum and classical channel uses~\cite{bowen2006stochastic}. Further, Bowen has shown that introducing additional noise during the decoding phase can improve the rate of classical information transmission, particularly with threshold detection methods~\cite{bowen2004noise}.

Recent applications of noise enhancement have extended to the realms of quantum state detection and quantum state estimation in scenarios involving noisy qubits~\cite{chapeau2015qubit, gillard2018qubit, gillard2017stochastic}. Studies indicate that representing the decohering environment as a thermal bath at a finite temperature and increasing this noise temperature can raise up the metrological performance of noisy qubits. While earlier investigations primarily focused on thermal noise with non-unital characteristics, Nicolas et al. have revealed that the introduction of unital noises can further refine the efficiency of phase estimation processes in noisy qubit environments~\cite{gillard2019stochastic}. Just as noise can subtly influence the behavior of quantum systems, it also plays a crucial role in the complex systems of neuroscience and biology. Here, the application of noise moves from the quantum scale to the macroscopic, affecting everything from neuron firing patterns to whole-body sensory and motor functions.

\subsubsection{Noise-Enhanced Neuroscience and Biology.}

The exploration of noise's constructive impact in neuroscience began with Bulsara's investigation into sensory information transmission within neuron models in 1991~\cite{bulsara1991stochastic, longtin1991time}. Recognizing that sensory systems often interpret weak signals, Douglass et al. applied external noise to crayfish mechanoreceptors in 1993, showcasing the stochastic resonance (SR) effect's enhancement of weak mechanical stimulus transmission~\cite{douglass1993noise}. Subsequent research expanded our understanding of noise's benefits in neuronal models~\cite{longtin1993stochastic, bulsara1996threshold,longtin1994bistability} and provided experimental evidence of SR in neurons within the cricket's cercal sensory system, triggered by externally applied noisy air currents~\cite{levin1996broadband}. A comprehensive review of these advancements in sensory neuron research is available in~\cite{moss2004stochastic}.

Moreover, noise's advantages extend to human health and bodily functions. As humans age, somatosensory capabilities diminish, potentially impairing motor control and increasing fall risk. Research on noise-enhanced balance control in humans has illustrated noise's capability to (1) augment motor control~\cite{priplata2002noise}, (2) enhance detection and signal transmission within the sensorimotor system during motor tasks~\cite{mendez2012improved}, (3) boost balance performance in older adults during visual sensory conflict tasks~\cite{dettmer2015effects}, and (4) heighten vibrotactile sensitivity in older adults and stroke patients~\cite{liu2002noise}. SR phenomena have been observed across all sensory modalities, including the visual system where noise improves contrast detection sensitivity and visual motion discrimination in healthy adults~\cite{trevino2016noise}, and the auditory system, aiding in the perception, detection, and discrimination of pure tones~\cite{zeng2000human, van2016transcranial}.

Despite the growing body of experimental~\cite{stacey2001synaptic,kole2006single} and theoretical~\cite{dykman1998can,hanggi2002stochastic} research on stochastic resonance within biology and neuroscience, some efforts focus predominantly on classical stochastic resonance. These studies often do not fully integrate with contemporary theoretical advancements or may overlook the biological relevance of critical metrics (e.g., signal, noise, and processing aspects) by adopting abstract models~\cite{mcdonnell2011benefits}. McDonnell et al. introduced the term ``stochastic facilitation" to more accurately describe noise's beneficial role in biological research, defining it as the improvement of a specific neural system's intended computational goal through the presence of stochastic biological noise~\cite{mcdonnell2011benefits}. 


\subsection{Benefits of Noise in Optimization Techniques}

Optimization methods can significantly benefit from the strategic use of randomness. Traditional random search optimization techniques often end up being trapped in local minima. In such scenarios, introducing random noise aids in initially exploring broader regions of the state space, thereby enhancing the likelihood of identifying optimal or near-optimal solutions. For instance, the integration of randomization within the crossover and mutation phases of Genetic Algorithms (GA)~\cite{tang1996genetic} prevents population self-similarity and circumvents local minima. Here, the mutation process, akin to noise injection, can, with an optimal mutation rate, boost algorithm performance~\cite{chen2014noise}.

\textit{Simulated Annealing (SA)}~\cite{science.220.4598.671}, another method geared toward approximating global optima, benefits from controlled randomness to facilitate convergence, proving particularly effective for combinatorial challenges such as graph partitioning. SA manipulates a temperature parameter $T$, gradually decreasing from an initial high value to zero. Higher values of $T$ increase the probability of accepting suboptimal solutions, a feature of randomness or "noise" within the algorithm~\cite{chen2014noise}. As $T$ diminishes, so does the noise variance, vanishing altogether when $T = 0$, and with it, the randomness, enabling the algorithm to reach global optimum.

\textit{Boltzmann machines} utilize a similar approach, employing noise temperature to navigate the state space efficiently~\cite{hinton1986learning}. At elevated noise levels, these machines can survey the entire state space structure, pinpointing a coarse minimum. The subsequent reduction in noise temperature allows for finer optimization within this broad minimum, aiming for the best achievable solution (optimal or near-optimal).

\subsection{Noise in Privacy Preservation}

Within the domain of privacy preservation, we navigate through two important strategies: Noise in Microdata Protection and Noise in Differential Privacy, showcasing how each method leverages noise to safeguard sensitive data while maintaining its analytical value.

\subsubsection{Noise in Microdata Protection.} Protecting individual respondent information in microdata, collected through surveys, is crucial. Masking techniques transform original data to safeguard respondent confidentiality while maintaining statistical validity~\cite{brand2002microdata, domingo2004security}. Perturbative masking, such as random noise, involves altering sensitive attributes by adding or multiplying them with random variables of a specified distribution~\cite{dalenius1977towards, ciriani2007microdata}. Considering $N$ tuples with $X_j$ as the sensitive attribute column, the uncorrelated additive random noise method substitutes each observation $x_{ij}$ with $x_{ij} + \epsilon_{ij}$, where $\epsilon_j \sim \mathcal{N}(0, \sigma_{\epsilon_j}^2)$ and $\sigma_{\epsilon_j}^2 = \alpha \sigma_{X_j}^2$, altering means while retaining covariances. Conversely, correlated noise addition ensures the error covariance matrix, $\epsilon \sim \mathcal{N}(0,\Sigma_\epsilon)$ where $\Sigma_\epsilon = \alpha\Sigma$, mirrors that of the original data, preserving correlation coefficients.

\subsubsection{Noise in Differential Privacy.}
Differential privacy benefits from noise enhancement to minimize single record identification risks in database releases~\cite{10.1007/11787006_1, dwork2008differential}. This approach ensures that the presence or absence of an individual's data in a dataset does not significantly affect the outcome of analyses performed on that dataset.
Achieving $\epsilon$-differential privacy or ($\epsilon$, $\delta$)-differential privacy involves adding random noise to the data or query results. The amount of noise added is calibrated to the sensitivity of the query function and the desired privacy level. Two common mechanisms for implementing differential privacy are the Laplace mechanism and the Gaussian mechanism.
The Laplace mechanism, used for $\epsilon$-differential privacy, introduces Laplace-distributed noise to data outputs. This is formalized in the following theorem:

\begin{thm}[$\epsilon$-Differential Privacy Theorem~\cite{dwork2006calibrating}] For any function $f : D \rightarrow R^k$ with sensitivity $\Delta f$, the mechanism $K_f$ adding noise from Lap($\frac{\Delta f}{\epsilon}$) to each output term upholds $\epsilon$-differential privacy for all datasets $D_1$, $D_2$ differing by at most one element. 
\end{thm}

Here, the sensitivity $\Delta f$ represents the maximum change in the function's output when a single record in the dataset is modified. 
For scenarios requiring a relaxation of privacy guarantees, ($\epsilon$, $\delta$)-differential privacy is often used. This approach employs the Gaussian mechanism, which incorporates Gaussian noise. The theorem for this mechanism is as follows:

\begin{thm}
[($\epsilon$, $\delta$)-Differential Privacy Theorem~\cite{dwork2014algorithmic}] Given $\epsilon \in (0, 1)$ and the $l_2$-sensitivity of $f : N^|x|\rightarrow R^k$ as $\Delta_2(f)$, the Gaussian mechanism with parameter $\sigma\geq \frac {c\Delta_2(f)}{\epsilon}$ ensures ($\epsilon$, $\delta$)-differential privacy for $c^2 > 2 \ln(\frac{1.25}{\delta})$.
\end{thm}

In this case, $\delta$ represents a small probability of privacy violation beyond the $\epsilon$ guarantee. The $l_2$-sensitivity $\Delta_2(f)$ measures the maximum change in the $l_2$ norm of the function's output when a single input record is changed.
These noise-based mechanisms demonstrate how controlled randomness can enhance privacy protection in data analysis and release. By carefully calibrating the noise to the sensitivity of the query and the desired privacy level, differential privacy provides a robust framework for balancing data utility with individual privacy. He et al. established a necessary and sufficient condition for $\epsilon$-differential privacy, as well as sufficient conditions for
($\epsilon$, $\delta$)-differential privacy~\cite{he2020differential}. They showed that the probability density function of noise should cover all possible numerical values that the noise can take, ensuring that it can impact any potential output from the system. Additionally, the ratio between the probabilities of different outputs (after noise is added) must stay within a certain limit. For further details on differential privacy, including various noise addition conditions, motivations, and applications, refer to He et al.~\cite{he2020differential} and Dwork et al.~\cite{dwork2008differential}.

\subsubsection{ Privacy in Graph Data.} Graph perturbation techniques, which include adding or removing edges and nodes, play a crucial role in the context of data privacy for graphs. Torra et al. introduced a new approach to graph anonymization by treating graph perturbation as noise addition~\cite{torra2019graph}, similar to methods used in database privacy. Torra et al. formalized this problem with definitions of "graph addition" and "noise graph addition," aligning nodes between original and noise graphs. This perspective improves understanding of privacy in social networks, balancing data protection with minimal information loss.

\section{Future Directions}
As discussed, the benefits of noise have been observed in different scientific domains. Building upon this foundation, there is significant potential for future research that looks deeper into both the theoretical and practical aspects of noise benefits. Here, we will present several key areas where further research can make significant contributions:\vspace{3mm}

\noindent\textbf{-- Theoretical and Applied Strategies for Utilizing Noise.} Similar to how Section~\ref{section:NNOutputNeuron} details the conditions under which noise enhances performance in regression and classification tasks, there is a growing need to develop robust theoretical results that can predict when and how noise becomes beneficial. By investigating the impact of different types of noises on algorithms, particularly in machine learning, we can better understand how specific forms of noise might improve model performance or uncover new insights~\cite{liu2022towards}. Key questions include: What is the best or optimal noise? What type of noise, and how much noise, should be introduced to achieve the desired enhancement in performance? Addressing these questions will enable us to systematically harness noise for better computational results.\vspace{2mm}

\noindent\textbf{-- Supervised Learning and Embedding Techniques.} Future research could explore noise injection strategies specifically tailored for supervised learning settings (see Section ~\ref{section:DL}) and embeddings, two well-known areas that are less explored for enhancement via noise injection. Investigating the impact of noise on the quality and stability of embeddings could be particularly insightful for applications in language models and recommendation systems.\vspace{2mm}

\noindent\textbf{-- Integration with Large Language Models (LLMs).} As large language models continue to grow in capability and application, exploring how noise interacts with these systems during both training and inference will reveal important future research directions. As we discussed in Section~\ref{section:NLP}, when noise is strategically introduced, it can enhance the robustness, adaptability, and human-like qualities of LLMs. Future research should aim to improve how noise is used during training to better mimic human learning, making models more reliable and accurate.\vspace{2mm}

\noindent\textbf{-- Synthetic Data Generation.} Noise is valuable for generating synthetic data that can be used to train machine learning models, especially in domains where real data is difficult to collect or restricted due to privacy concerns~\cite{fonseca2023tabular}. Adaptive noise injection techniques, which adjust noise based on domain-specific features, have emerged as an effective approach to enhance the realism and utility of synthetic datasets~\cite{10.1016/j.cviu.2023.103855}. Furthermore, the combination of synthetic data generation with privacy-preserving methods, like differential privacy, is becoming more important. Techniques such as PrivBayes~\cite{zhang2017privbayes} and PATE-GAN~\cite{jordon2018pate} demonstrate how noise can protect sensitive information while maintaining the usefulness of the data. Future research could focus on integrating adaptive noise injection with differential privacy to improve both data realism and security~\cite{fonseca2023tabular}.\vspace{2mm}

\noindent\textbf{-- Image Conditional Generation.} Conditional generation is creating outputs based on specific input conditions or constraints. Future research in conditional generation tasks, such as super-resolution and image synthesis, could benefit from noise. The Noise Conditional Flow Model for Super-Resolution (NCSR) introduces noise to enhance diversity and visual quality in generated images~\cite{kim2021noise}. NCSR has shown that adding noise to the training data and incorporating a noise conditional layer can overcome issues like data distribution mismatch and improve the generation of multiple high-resolution images from a single low-resolution input. Integrating such methods could pave the way for future advances in conditional generation, offering better control over complexity and variability in generated outputs.\vspace{2mm}

\noindent\textbf{-- Graph Neural Networks.} As previously discussed in Section~\ref{section:graphAugmentation}, future research should continue exploring the use of noise as a regularization tool in GNNs. Previous studies, like FLAG~\cite{kong2022robust}, have shown that adding adversarial noise during training can significantly enhance GNN performance across multiple tasks. Similarly, Sato et al. (2021) demonstrated that random input features improve generalization in GNNs~\cite{sato2021random}. Looking ahead, techniques like ``Noisy Nodes," which introduce noise to the input graph and apply a noise-correcting loss at the node level, show promise in developing diverse and meaningful node representations~\cite{godwin2021simple}. These findings suggest that noise-based regularization techniques could play a key role in enhancing generalization and robustness in future GNN applications.

\section{Conclusion}
We have reviewed the multifaceted landscape of noise-enhanced systems and techniques across various domains, from systems and learning algorithms to specialized applications in graph machine learning and natural language processing. The strategic injection of noise, once considered merely a source of error or unwanted variability, has been reimagined as a powerful tool for enhancing signal detection, learning efficacy, privacy preservation, and even the robustness of natural language models. Through a careful examination of noise-enhanced methodologies, we have seen how introducing randomness can paradoxically lead to greater stability, and accuracy in the extraction of information from data. This survey emphasized the importance of noise not as a barrier, but as a helper in our search for deeper understanding and new ideas in the big data era.

\bibliographystyle{ACM-Reference-Format}
\bibliography{main}


\end{document}